\documentclass[conference]{IEEEtran}
\IEEEoverridecommandlockouts

\usepackage{cmap}        
\usepackage[T1]{fontenc} 
\usepackage{hyperref}

\usepackage{orcidlink}
\usepackage{graphicx}%
\usepackage{multirow}%
\usepackage{amsmath,amssymb,amsfonts}%
\usepackage{amsthm}%
\usepackage{mathrsfs}%
\usepackage{xcolor}%
\usepackage{xurl}%
\usepackage{textcomp}%
\usepackage{manyfoot}%
\usepackage{booktabs}%
\usepackage{algorithm}%
\usepackage{listings}%
\usepackage{enumitem}
\usepackage{xurl}
\usepackage{tabularx}
\usepackage{bbding}
\usepackage{array}

\usepackage{cite}
\usepackage{amsmath,amssymb,amsfonts}
\usepackage{algorithmic}
\usepackage{graphicx}
\usepackage{textcomp}
\usepackage{xcolor}
\def\BibTeX{{\rm B\kern-.05em{\sc i\kern-.025em b}\kern-.08em
    T\kern-.1667em\lower.7ex\hbox{E}\kern-.125emX}}

\makeatletter
\newcommand{\linebreakand}{%
  \end{@IEEEauthorhalign}
  \hfill\mbox{}\par
  \mbox{}\hfill\begin{@IEEEauthorhalign}
}
\makeatother
    
\begin{document}

\title{Evaluating and Enhancing LLMs for
Multi-turn Text-to-SQL with Multiple Question Types\thanks{This work was supported by the National Natural Science Foundation of China (Grant No. 62172123) and the Key Research and Development Program of Heilongjiang Province (Grant No. 2022ZX01A36).}}

\author{
	\IEEEauthorblockN{
		Ziming Guo\orcidlink{0009-0000-6439-3219}\IEEEauthorrefmark{1}, 
		Chao Ma\IEEEauthorrefmark{1}, 
		Yinggang Sun\IEEEauthorrefmark{2}, 
		Tiancheng Zhao\IEEEauthorrefmark{1} 
		Guangyao Wang\orcidlink{0009-0003-7671-6195}\IEEEauthorrefmark{1}
            Hai Huang\IEEEauthorrefmark{1}} 
\IEEEauthorblockA{\IEEEauthorrefmark{1}School of Computer Science, Harbin University of Science and Technology\\ Harbin, China 150040\\ Email: 2204050108@stu.hrbust.edu.cn, machao8396@163.com,\\ 1783467143@qq.com, 2204050124@stu.hrbust.edu.cn, hust\_hh@vip.163.com}
	\IEEEauthorblockA{\IEEEauthorrefmark{2} Faculty of Computing, Harbin Institute of Technology, Harbin, China 150001\\ Email: 23b903085@stu.hit.edu.cn}} 

\maketitle
\begin{abstract}
Recent advancements in large language models (LLMs) have significantly improved text-to-SQL systems. However, most datasets and LLM-based methods tend to focus narrowly on SQL generation, often neglecting the complexities inherent in real-world conversational queries. This oversight can result in unreliable responses, particularly for ambiguous questions that cannot be directly addressed with SQL. To address this gap, we propose MMSQL, a comprehensive test suite designed to evaluate LLMs' question classification and SQL generation capabilities by simulating real-world scenarios with diverse question types and multi-turn Q\&A interactions. Utilizing MMSQL, we assessed the performance of popular LLMs, including both open-source and closed-source models, and identified key factors influencing their performance in these contexts. Furthermore, we introduce an LLM-based multi-agent framework that employs specialized agents to identify question types and determine appropriate answering strategies. Experimental results demonstrate that this method effectively enhances baseline models' ability to handle diverse question types in conversational scenarios. Our approach simultaneously considers multiple question types and multi-turn interactions, providing a new, realistic perspective, and offering a valuable advancement toward more reliable and versatile text-to-SQL systems. Our dataset and code are publicly available at \href{https://mcxiaoxiao.github.io/MMSQL}{https://mcxiaoxiao.github.io/MMSQL}.
\end{abstract}


\section{Introduction}\label{sec1}

Text-to-SQL bridges natural language and SQL queries, empowering non-technical users to access and analyze data without mastering complex SQL knowledge. The advent of LLMs, with their remarkable capacity for following instructions, has transformed the text-to-SQL domain. LLM-based methods have delivered remarkable outcomes across various text-to-SQL tasks \cite{xiong2024interactivet2smultiturninteractionstexttosql,lian2024chatbinaturallanguagecomplex}, while the assessment of their robustness is increasingly drawing attention. Current research often assumes that user queries are unambiguous, focusing on expanding the scope of databases and increasing the complexity of questions to elevate the difficulty of question-to-SQL mapping, thereby pushing the boundaries of text-to-SQL systems. \cite{li2024can,wang2024conda,pourreza2024chasesqlmultipathreasoningpreference,wang2024macsqlmultiagentcollaborativeframework}. However, in practical scenarios, user queries are inherently dynamic and uncertain \cite{wang2023know}. Assuming that all user queries can be answered may lead to the generation of hallucinated content and unreliable predictions, thereby undermining the dependability of these systems \cite{rawte2023surveyhallucinationlargefoundation,saparina2024}. 

\begin{figure*}[h]
  \centering
  \includegraphics[width=0.99\textwidth]{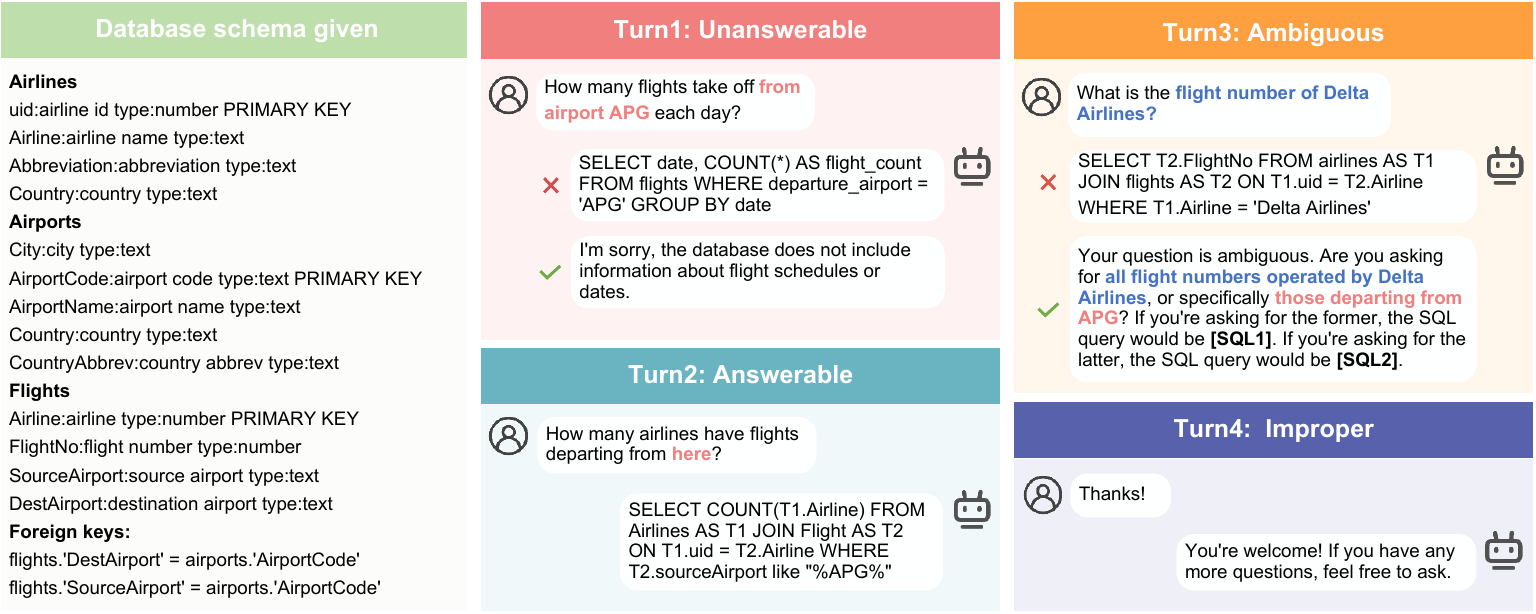}
  \label{fig:types}
  \caption{A four-turn dialogue example that includes four different types of Q\&A pairs.}
\end{figure*}

Observing real-world conversations, it is common for users to engage in multi-turn dialogues with interrelated questions that do not readily translate into SQL. In such scenarios, relying on single-turn or assumptive responses can lead to suboptimal performance, as it may not fully address the user's needs or context \cite{wang2023know,lee2024trustsql,he2023text2analysisbenchmarktablequestion}. Figure \ref{fig:types} illustrates examples of responses to three common types of such questions: 
(1) \textit{Unanswerable:} If the database lacks data, such as the flight departure times from an airport queried in turn 1, the system needs to explain its inability to provide an answer.
(2) \textit{Ambiguous:} In turn 3, the query about the flight number of Delta Airlines might refer to conditions from turn 1 and turn 2: "from airport APG", or it might simply mean all flight numbers operated by Delta Airlines. The system needs to recognize this ambiguity and respond accordingly to avoid providing incorrect SQL responses.
(3) \textit{Improper:} Questions unrelated to the database content, such as casual conversation in turn 4, should not generate SQL-based responses. These examples highlight the necessity for reliable models capable of navigating various types of complex conversational dynamics. There is a need for more robust models that can effectively manage such challenges.

Existing studies have only sporadically addressed the effective assessment and handling of multi-type questions. Most text-to-SQL research focuses on achieving high accuracy for single-type or single-round user questions \cite{wang2024conda, wang2024macsqlmultiagentcollaborativeframework, yu2019sparc, guo2021chase}, often overlooking the need to develop systems capable of multi-turn dialogues that handle a variety of question types \cite{li2024can}. This limited scope can result in incorrect responses, especially for those questions that do not necessitate SQL solutions. It has been observed that even the most sophisticated LLMs struggle to effectively address these challenges \cite{saparina2024}. Although recent studies have acknowledged the challenges of ambiguous and unanswerable questions \cite{wang2023know}. However, existing solutions often resort to simple abstention when faced with vague or conversational elements, lacking granular designs to manage various question types, leading to passive model responses and hindering practical applicability and robustness \cite{lee2024trustsql, zhang2020did, sun2024qda}. In summary, while individual studies have made some progress in evaluating and enhancing the reliability of text-to-SQL systems, these efforts fall short of a comprehensive solution to the problem.

In response to these challenges, we introduce the Multi-type and Multi-turn text-to-SQL test suite (MMSQL), a comprehensive benchmark engineered to evaluate the proficiency of LLMs in handling multi-turn text-to-SQL tasks across diverse question types. Utilizing MMSQL, we have undertaken an exhaustive performance analysis of leading LLMs, encompassing open-source and closed-source models. This analysis provides detailed insights, highlighting the promise of open-source models and  reveals the suboptimal performance of models on specific question types and the varying effectiveness of strategies designed to resolve ambiguity. As an initial attempt to address these challenges, we also propose an innovative LLM-based multi-agent framework, inspired by top-performing models and informed by these insights. This framework is anchored by a core Question Detector and Question Decomposer, tasked with identifying question types and determining appropriate answering strategies. It could provide potential SQL queries for ambiguous questions, accounting for multiple interpretations. Furthermore, the framework includes two supportive agents: the Schema Selector, which identifies and provides the essential subset of a database schema, and the SQL Refiner, which is dedicated to refining SQL queries. Our experiments, conducted on the MMSQL benchmark, have shown that this framework yields a significant performance improvement.

Our main contributions are summarized as follows:

\begin{enumerate}
\item We developed MMSQL, a comprehensive test suite designed to evaluate the performance of LLMs in multi-turn text-to-SQL conversations across diverse question types.

\item We provide an in-depth analysis of popular LLMs' performance on MMSQL, offering insights into their capabilities and identifying key factors that drive performance differences when handling a variety of question types.

\item We propose a novel multi-agent framework adept at addressing text-to-SQL tasks for various question types, evaluated on MMSQL, and demonstrated to be effective through a series of ablation experiments.
\end{enumerate}

\section{Related Work}\label{sec2}
\subsection{Text-to-SQL}

Text-to-SQL involves transforming natural language queries into SQL, representing a crucial area of natural language processing with many practical applications. Essential datasets include single-turn datasets such as WikiSQL \cite{zhong2017seq2sqlgeneratingstructuredqueries}, Spider\cite{yu2018spider}, and BIRD \cite{li2024can}, as well as multi-turn datasets such as CHASE \cite{guo2021chase}, SParC \cite{yu2019sparc}, and CoSQL \cite{yu2019cosql}. These datasets serve as benchmarks for evaluating the performance of text-to-SQL systems. In recent years, the emergence of pre-trained language models (PLMs) has driven significant progress in this domain. Models like SQLova \cite{hwang2019comprehensiveexplorationwikisqltableaware}, PICARD \cite{scholak2021picard}, and RAT-SQL \cite{wang-etal-2020-rat} have achieved impressive results. However, despite their effectiveness, these specialized models require considerable refsources and time for training. LLMs have revolutionized numerous NLP tasks, including text-to-SQL, surpassing previous models and establishing a new paradigm \cite{sala2024text}. Recent studies have focused on enhancing prompt design and developing complex, multi-stage frameworks that leverage LLMs to enhance performance \cite{mostajabdaveh2024optimization}. Notably, LLM-based frameworks like DIN-SQL \cite{pourreza2023dinsqldecomposedincontextlearning}, which utilize chain-of-thought reasoning \cite{wei2022chain}, decompose the problem into simpler sub-problems. MAC-SQL \cite{wang2024macsqlmultiagentcollaborativeframework} and CHASE-SQL \cite{pourreza2024chasesqlmultipathreasoningpreference} employ multi-agent collaborative frameworks, demonstrating significant advancements in this field.

Recent studies highlight the inherent diversity and ambiguity of natural language, noting that not all user queries can be effectively translated into accurate SQL as the answer in practical applications \cite{wang2023know, lee2024trustsql,10598154}. Datasets such as TriageSQL \cite{zhang2020did}, NoisySP \cite{wang2023know}, and CoSQL have incorporated question type detection as part of the models' tasks. Previous work, TrustSQL \cite{lee2024trustsql} demands that models opt for abstention when faced with diverse unanswerable questions. AMBROSIA \cite{saparina2024} and \cite{bhaskar-etal-2023-benchmarking} propose testing text-to-SQL systems by parsing ambiguous questions into multiple potential SQL queries, thereby enhancing model usability in real-world scenarios. \cite{bhaskar-etal-2023-benchmarking} improves beam search on T5 \cite{raffel2023exploringlimitstransferlearning} to provide possible SQL responses for ambiguous questions and evaluate their effectiveness. Additionally, \cite{sun2024qda} employs questions enhanced dialogue augmentation to train open-source LLMs to identify question types. These studies focus solely on the model's ability to recognize question types or address specific problems, which may result in unhelpful abstentions or hallucinations due to specialized question-and-answer tailoring \cite{ye2023cognitive}. Moreover, most work is limited to single-turn settings, whereas real interactions often involve ambiguity that must be addressed in multi-turn contexts. These limitations motivate our work, which evaluates and enhances LLMs for multi-turn dialogues with multiple question types.

\subsection{LLM-based Agents}

LLM-based agents have long captured the attention of researchers in both academia and industry \cite{wang2023survey}. With the expansion of web knowledge, LLMs are increasingly demonstrating intelligence levels comparable to humans. This evolution has sparked a growing interest in creating autonomous agents driven by LLMs. AutoGPT \cite{Significant_Gravitas_AutoGPT} and MetaGPT \cite{hong2024metagptmetaprogrammingmultiagent} enhance AI models by integrating a range of useful tools, allowing developers to create adaptable, conversational agents that can function in various modes by leveraging LLMs, human input, and other tools to complete tasks. In the realm of text-to-SQL parsing, frameworks such as MAC-SQL \cite{wang2024macsqlmultiagentcollaborativeframework}, MAG-SQL \cite{xie2024magsqlmultiagentgenerativeapproach}, and CHASE-SQL \cite{pourreza2024chasesqlmultipathreasoningpreference} employ multiple LLM-based agents to interpret SQL queries collaboratively. This can typically be summarized as a three-step process: retrieving relevant values, generating SQL, and refining the SQL output. These multi-agent systems have achieved leading results on the BIRD benchmark \cite{li2024can}. However, existing methodologies have yet to incorporate pipeline designs specifically tailored to manage multiple types of questions concurrently within interactive environments. Drawing inspiration from these advancements, we developed a multi-agent framework incorporating a Question Detector Agent, designed to effectively manage the complexity and variety of user queries encountered in real-world applications.

\section{MMSQL: Multi-type and Multi-turn Text-to-SQL Test Suite}\label{sec3}

\begin{table*}[ht]
\caption{Comparison of multi-turn or multi-type text-to-SQL datasets.}
\centering
\begin{tabular}{lcccccc} 
\toprule
& SParC & CoSQL &NoisySP &AmbiQT &AMBROSIA & MMSQL \\ 
\midrule
\# Dialogues & 4,298 & 3,007 & - & - & - & \textbf{6493} \\ 
Total \# turns & 12,726 & 15,433 & 15,598 &3,046 & 4,242 & \textbf{38,666} \\ 
Avg. \# Q turns & 3.0  & 5.2 & 1 & 1 & 1 & \textbf{6.0} \\ 
Avg. Q len & 10.2 & 11.2 & - & - & - & \textbf{11.4} \\ 
Ans. Q type & \Checkmark  & \Checkmark & \XSolidBrush & \XSolidBrush & \Checkmark  & \Checkmark \\ 
Amb. Q type & \XSolidBrush  & \Checkmark & \Checkmark & \Checkmark & \Checkmark & \Checkmark \\ 
Una. Q type & \XSolidBrush  & \Checkmark & \Checkmark & \XSolidBrush  & \XSolidBrush & \Checkmark \\ 
Imp. Q type & \XSolidBrush  & \Checkmark & \XSolidBrush & \XSolidBrush & \XSolidBrush &  \Checkmark \\ 
\bottomrule
\end{tabular}
\footnotetext{The numbers are computed based on the total number of user utterances in the train set (if included), consistent with previous works. SParC \cite{yu2019sparc} and CoSQL \cite{yu2019cosql} are multi-turn datasets, while NoisySP \cite{wang2023know}, AmbiQT \cite{bhaskar-etal-2023-benchmarking} and AMBROSIA \cite{saparina2024} are single-turn datasets.}
\label{table:comparison}
\end{table*}

\subsection{Task Formulation}
MMSQL aims to assess the effectiveness of language models in managing multi-type and multi-turn text-to-SQL tasks. By analyzing real-world text-to-SQL interactions, we have categorized four types of question-answer scenarios, as shown in Figure \ref{fig:types}.

\begin{enumerate}

\item \textbf{Unanswerable}:
These questions pertain to information that is not present in the database or exceeds the system's operational scope, such as when they involve external knowledge or web search capabilities beyond the SQL system. The system should acknowledge its inability to answer the question and explain the reason to the user. For example, a question about the airport's flight schedule each day, which is not present in the database, should be identified by the system as unanswerable.

\item \textbf{Answerable}:
These questions can be directly answered using SQL queries based on the available database information. The system must generate an accurate SQL query reflecting the user's request and execute it to retrieve the required data. For example, in Turn 2, when asked, "How many airlines have flights departing from here?" the system correctly generates the SQL query by interpreting "here" within the given context to find airlines with flights departing from the specified location.

\item \textbf{Ambiguous}:
These questions include terms that might correspond to multiple columns or values in the database, causing potential ambiguity in SQL query generation. In a multi-turn context, the system should identify and address this ambiguity by seeking clarification from the user. For example, in Turn 3, when asked, "What is the flight number of Delta Airlines?" the system should recognize the potential ambiguity—whether the user is asking for all flights or specific ones from a location like APG. It should ask for clarification (e.g., "Are you asking about all flights or specific ones?") and simultaneously propose possible SQL queries based on these interpretations. This approach ensures the system can adapt and refine its responses through ongoing dialogue.

\item \textbf{Improper}:
These questions are irrelevant to the database or pertain to everyday conversation that does not require an SQL response. The system should recognize these questions and respond appropriately, without attempting to generate an SQL query. For example, a casual "thank you" from the user should be met with a simple "You're welcome!" rather than an SQL statement.
\end{enumerate}

In this task, the model is required to perform two key functions: first, to accurately determine the type of each question; and second, to attempt resolution and provide answers for questions categorized as ambiguous and answerable. This dual focus ensures that the model classifies questions effectively and actively generates appropriate SQL queries for complex scenarios. For questions deemed improper or unanswerable, the model aims to provide the most helpful natural language responses possible.

\subsection{Construction}

The MMSQL data comes from two sources: refined samples from CoSQL and new samples generated by QDA-SQL (Questions Enhanced Dialogue Augmentation for Multi-turn Text-to-SQL) \cite{sun2024qda}. QDA-SQL uses Chain of Thought (CoT) \cite{wei2022chain} to guide new sample generation through step-by-step reasoning. This involves multi-turn Q\&A generation, and LLM-based refinement to help Gemini-Pro generate diverse samples. Each sample aligns with predefined question types. Appendix \ref{QDA-SQL} provides further details on its effectiveness and design.

\begin{figure}[h]
    \centering
    \includegraphics[width=0.99\linewidth]{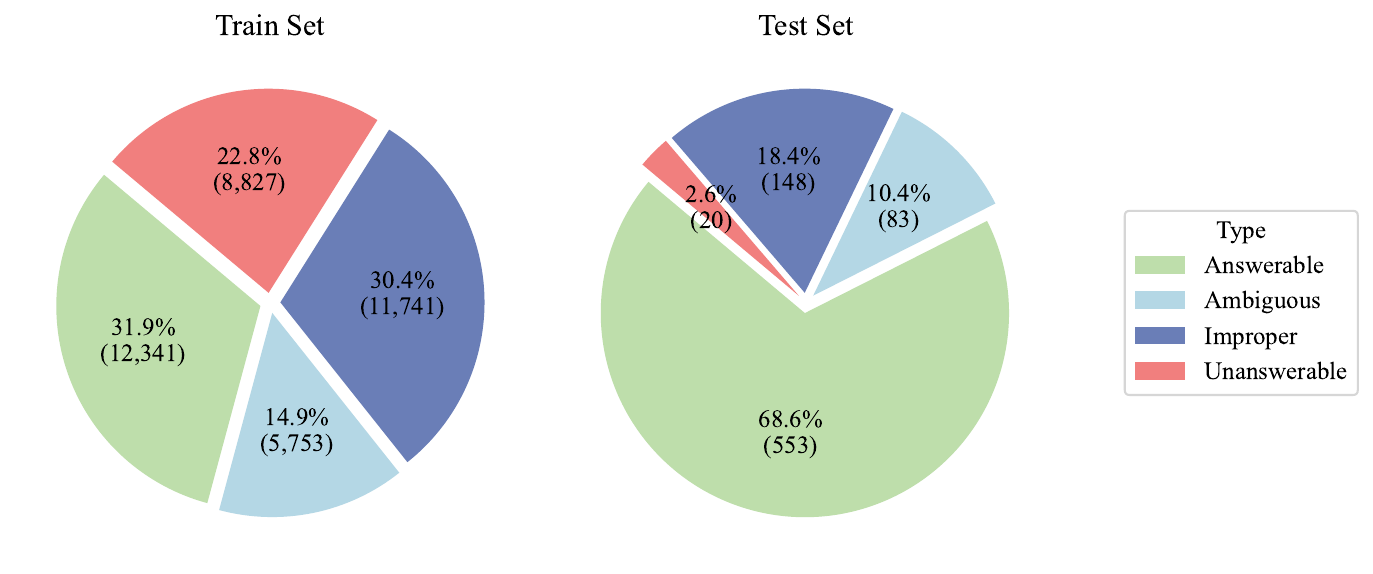}
    \caption{Distribution of question types in the curated MMSQL dataset, highlighting the proportions of Answerable, Ambiguous, Improper, and Unanswerable questions in both the training set and the test set.}
    \label{fig:typepie}
\end{figure}

We carefully screened initial samples to ensure high standards of difficulty, classification accuracy, and diversity, resulting in a refined dataset. Modifications transformed some instances into a curated MMSQL test set, with a focus on clearly distinguishing unanswerable and ambiguous questions during manual annotation. The dataset includes 6,493 training rounds and 149 test rounds of dialogue. Figure \ref{fig:typepie} shows the distribution of question types, highlighting that the training set covers all four types sufficiently, while the test set reflects a more realistic distribution \cite{wang2023know}. Table \ref{table:comparison} compares MMSQL with other multi-turn or multi-type datasets, MMSQL uniquely includes four question types and offers more dialogues with a higher average number of turns per dialogue, providing extensive multi-turn training data across diverse query types.

\subsection{Evaluation Metrics}

Following BIRD \cite{li2024can} and CoSQL \cite{yu2019cosql}, our evaluation metrics include Exact Matching (EM) and Execution Accuracy (EX). EM compares each predicted clause to the reference query, considering it is correct only if all components match, excluding values. EX evaluates the proportion of SQL where the execution results of both predicted and ground-truth SQL are identical. Interaction Execution Accuracy (IEX) is achieved when all SQL queries in a multi-turn interaction execute correctly. We also developed the Dual Assessment of Question Type Detection and Execution Accuracy (TDEX) to assess the comprehensive capability of text-to-SQL models with complex queries. Additionally, the Response Quality Score (RQS) measures the quality of the model's natural language responses.

TDEX is a comprehensive metric that evaluates the accuracy of user query type classification and execution accuracy.  For a set of \( N \) questions, where \( C_i \) denotes the expected classification and \( \hat{C}_i \) represents the predicted classification for the \( i \)-th question, \( S_i \) denotes the ground truth SQL query and \( \hat{S}_i \) represents the predicted SQL query for the \( i \)-th question, TDEX is computed as:
\begin{equation}
  \label{eq:example3}
    \text{TDEX} = \frac{1}{N} \sum_{i=1}^{N} \left\{
    \begin{array}{ll}
    \varepsilon_{\text{exec}}(S_i, \widehat{S}_i) & \text{(a)} \\
    \delta_{\text{type}}(C_i, \widehat{C}_i) & \text{(b)}\\
    \end{array}
    \right.
\end{equation}

\begin{equation*}
\begin{array}{ll}

  \text{(a)} &  C_i = \mathrm{'Answerable' }  ~or~  \mathrm{ 'Ambiguous'} \\

  \text{(b)} & \mathrm{otherwise}

\end{array}
\end{equation*}
\noindent

where \( \varepsilon_{\text{exec}} = 1 \) if the execution result of \( \hat{S}_i \) matches the execution result of \( S_i \), and \( \varepsilon_{\text{exec}} = 0 \) otherwise; \( \delta_{\text{type}} = 1 \) if \( \hat{C}_i \) matches \( C_i \), and \( \delta_{\text{type}} = 0 \) otherwise.

To evaluate the quality of a model's responses, MMSQL utilizes an LLM-assisted rating method assessing Utility, Accuracy, Completeness, Clarity, and Relevance on a 0-to-2 scale, with a maximum score of 10. This approach, where LLMs directly score responses, has been demonstrated to be superior in assessing answer quality compared to traditional metrics like BLEU, ROUGE, and BERT-Score. \cite{zhou2023instructionfollowingevaluationlargelanguage,liu-etal-2023-g}. The detailed prompts used for this evaluation are provided in Appendix \ref{rqsp}. We employ GPT-4o-mini due to its strong alignment with human judgment \cite{kwan2024mt,zheng2024judging} and enhance evaluation calibration using the Multiple Evidence Calibration methodology \cite{wang2023largelanguagemodelsfair}. To ensure alignment with human judgment, three experts scored each of 100 samples, and their scores were averaged. Pearson, Spearman, and Kendall's Tau correlations were calculated, confirming a strong positive correlation between LLM and human scores, as shown in Table \ref{table:correlation}, demonstrating its reliability.

\begin{table}[h]
\centering

\caption{Correlation analysis between human and GPT-4 ratings for responses to different types of questions where SQL answer is not applicable.}
\begin{tabular}{>{\centering\arraybackslash}p{0.5cm}|>
{\centering\arraybackslash}p{0.7cm}>{\centering\arraybackslash}p{1cm}|>{\centering\arraybackslash}p{0.7cm}>
{\centering\arraybackslash}p{1cm}|>{\centering\arraybackslash}p{0.7cm}>{\centering\arraybackslash}p{1cm}}
\toprule
Type       & Pearson & P-value & Spearman & P-value & Kendall & P-value \\ 
\midrule
Una. & 0.86                & 5.4e-07                & 0.57                 & 7.6e-03                & 0.54               & 8.7e-03                \\
Amb. & 0.96               & 4.0e-08                & 0.88                & 3.0e-05                & 0.81               & 3.8e-04                 \\
Imp.      & 0.71                & 3.4e-23                & 0.62                & 1.0e-16               & 0.61              & 7.1e-14              \\ 
\bottomrule
\end{tabular}
\label{table:correlation}
\end{table}


\subsection{Baselines}
We evaluate eight popular LLMs' performance, including close-source and open-source options. The close-source LLMs evaluated are GPT-4 Turbo\footnote{\url{https://platform.openai.com/docs/models/GPT-4 Turbo-and-gpt-4}}, GPT-3.5 Turbo\footnote{\url{https://platform.openai.com/docs/models/gpt-3-5-turbo}}, and Gemini-1.5 Flash\footnote{\url{https://deepmind.google/technologies/gemini}}. The open-source LLMs include Llama-3 (70B and 8B)\footnote{\url{https://llama.meta.com/llama3}}, Llama-3-SQLCoder-8B\footnote{\url{https://defog.ai}} (a Llama-3 model specifically trained in the text-to-SQL domain), Codellama-7B\footnote{\url{https://www.llama.com/docs/integration-guides/meta-code-llama}}, and Mistral-7B-v0.2\footnote{\url{https://mistral.ai}}.

\subsection{Result of MMSQL Evaluation}
\label{sec:resultmmsql}

Table \ref{table:expres} and Table \ref{table:maintype} showcases the performance of the evaluated LLMs in a zero-shot setting on the multi-turn text-to-SQL task, covering four types of questions from the MMSQL test set. We assessed the models based on overall performance (TDEX), SQL generation performance (EX and EM), and their capability to recognize different question types. A detailed discussion of the results is provided in the following sections.

\begin{table}[h]
\centering

\caption{Performance metrics of the models on the MMSQL test set, with all values expressed in percentages, except for the RQS. }
\begin{tabular}{cccccc}
\toprule
\multirow{2}{*}{Model} & \multirow{2}{*}{TDEX} & \multirow{2}{*}{IEX} & \multirow{2}{*}{EX} & \multirow{2}{*}{EM} & \multirow{2}{*}{RQS} \\
                 &               &               &               &               &               \\ \hline
GPT-4 Turbo      & \textbf{67.0} & \textbf{30.2} & \textbf{70.0} & {\underline{}51.0}    & \textbf{5.80} \\
GPT-3.5 Turbo    & 64.1          & 25.5          & 69.6          & 47.3          & 4.74          \\
Gemini-1.5 Flash & {\underline{65.8}}    & {\underline{30.1}}    & \textbf{70.0} & \textbf{52.3} & 4.03          \\  \midrule
Llama3-70B       & 62.8          & 22.8          & 66.4          & 47.4          & 3.86          \\
Llama3-8B        & 64.0          & 20.1          & 66.1          & 45.7          & 4.55          \\
SQLCoder-8B      & 30.7          & 24.8          & 63.2          & 31.0          & 3.43          \\
Codellama-7B     & 30.7          & 3.4           & 27.2          & 21.7          & {\underline{5.09}}    \\
Mistral-7B-v0.2  & 26.4          & 0.7           & 13.7          & 11.2          & 4.37          \\ \bottomrule
\end{tabular}
\label{table:expres}
\end{table}

\paragraph{Comparative Performance of Closed-Source and Open-Source LLMs}

As shown in Table \ref{table:expres}, GPT-4 Turbo demonstrated exceptional performance with a TDEX score of 67.0, while Gemini-1.5 Flash scored 65.8. GPT-4 Turbo also achieved the highest RQS of 5.8, indicating its strong overall capabilities. Meanwhile, Llama3-70B showed comparable TDEX performance to GPT-3.5, scoring 62.8 and 64.1 respectively, highlighting the robust potential of closed-source models in zero-shot settings. Notably, Llama3-8B achieved a TDEX of 64.0, slightly surpassing GPT-3.5.

Regarding question type recognition, as shown in Table \ref{table:maintype}, GPT-4 achieved the highest average F1 score of 68.2. Open-source models also performed commendably, with Llama3-8B attaining an average F1 score of 64.2, demonstrating their competitive edge. While closed-source models exhibit superior performance across many metrics, open-source models are rapidly catching up.

\paragraph{Inferior Performance in Specific Question Types}

As shown in Table \ref{table:maintype}, both open-source and closed-source models face significant challenges with unanswerable and ambiguous questions, with the latter being particularly difficult. For example, despite demonstrating strong overall performance, GPT-4 Turbo, which achieves the highest question type F1 score of 68.2, exhibits much lower precision and recall for unanswerable questions (56.9 and 38.4) and ambiguous questions (25.9 and 70.0) compared to other types. This pattern is also consistent across other models, highlighting a widespread difficulty in accurately distinguishing such question types.

\begin{table}[h]
\centering
\caption{Response quality scores of models in handling different types of questions: Answerable (Ans.), Unanswerable (Una.), and Ambiguous (Amb.), along with the average score.}
\begin{tabular}{ccccl}
\toprule
Model                & \multicolumn{1}{l}{Una.} & \multicolumn{1}{l}{Amb.} & \multicolumn{1}{l}{Imp.} & Ave. \\ \midrule
GPT-4 Turbo                        & {\underline{9.38} }                               & \textbf{7.00}                          & \textbf{9.43}                         & \textbf{8.60}    \\
{GPT-3.5 Turbo} & 8.38                                      & {\underline{4.73} }                            & 8.29                                  & 7.13             \\
Gemini-1.5 Flash                     & \textbf{9.78}                             & 4.31                                   & {\underline{8.94} }                            & {\underline{7.68}}       \\   \midrule
Llama3-70B                     & 8.80                                      & 4.60                                   & 8.38                                  & 7.26             \\
Codellama-7B                   & 4.75                                      & 3.57                                   & 8.31                                  & 5.54             \\ \bottomrule
\end{tabular}
\footnotetext{The best score in each column is highlighted in \textbf{bold}, and the second-highest score is \underline{underlined}.}
\label{table:type_analysis}
\end{table}

In terms of natural language response quality, as shown in Table \ref{table:type_analysis}, Gemini-1.5 Flash performs well on unanswerable questions with a score of 9.78, but significantly drops to 4.31 on ambiguous questions. Codellama-7B is the lowest-performing model overall, with an average score of 5.54, particularly struggling with ambiguous questions at a score of only 3.57. This trend underscores a common challenge all models face in interpreting and responding to ambiguous queries, indicating a need for further development in handling ambiguity.

\begin{table*}[h]
\centering
\caption{Detailed performance metrics of the models on the MMSQL test set, with all values presented in percentages, presenting precision and recall for specific question types, along with Macro-F1 scores.}

\begin{tabular}{cclclclclclclclclcl}
\toprule
\multirow{2}{*}{Model} &
  \multicolumn{4}{c}{Ans.} &
  \multicolumn{4}{c}{Una.} &
  \multicolumn{4}{c}{Amb.} &
  \multicolumn{4}{c}{Imp.} &
  \multirow{2}{*}{ ~F1} &
  \multicolumn{1}{l}{} \\
 &
  \multicolumn{2}{c}{Prec} &
  \multicolumn{2}{c}{Rec} &
  \multicolumn{2}{c}{Prec} &
  \multicolumn{2}{c}{Rec} &
  \multicolumn{2}{c}{Prec} &
  \multicolumn{2}{c}{Rec} &
  \multicolumn{2}{c}{Prec} &
  \multicolumn{2}{c}{Rec} &
   &
  \multicolumn{1}{l}{} \\ \hline
GPT-4 Turbo &
  \multicolumn{2}{c}{{\underline{90.3}}} &
  \multicolumn{2}{c}{89.8} &
  \multicolumn{2}{c}{25.9} &
  \multicolumn{2}{c}{{\underline{70.0}}} &
  \multicolumn{2}{c}{56.9} &
  \multicolumn{2}{c}{38.4} &
  \multicolumn{2}{c}{\textbf{100.0}} &
  \multicolumn{2}{c}{{\underline{98.0}}} &
  \multicolumn{2}{c}{\textbf{68.2}} \\
GPT-3.5 Turbo &
  \multicolumn{2}{c}{88.8} &
  \multicolumn{2}{c}{89.2} &
  \multicolumn{2}{c}{16.0} &
  \multicolumn{2}{c}{65.0} &
  \multicolumn{2}{c}{64.3} &
  \multicolumn{2}{c}{20.9} &
  \multicolumn{2}{c}{\textbf{100.0}} &
  \multicolumn{2}{c}{96.0} &
  \multicolumn{2}{c}{61.1} \\
Gemini-1.5 Flash &
  \multicolumn{2}{c}{85.9} &
  \multicolumn{2}{c}{{\underline{95.8}}} &
  \multicolumn{2}{c}{26.3} &
  \multicolumn{2}{c}{50.0} &
  \multicolumn{2}{c}{58.3} &
  \multicolumn{2}{c}{8.1} &
  \multicolumn{2}{c}{\textbf{100.0}} &
  \multicolumn{2}{c}{95.4} &
  \multicolumn{2}{c}{59.3} \\ \midrule
Llama3-70B &
  \multicolumn{2}{c}{84.9} &
  \multicolumn{2}{c}{95.2} &
  \multicolumn{2}{c}{{\underline{27.5}}} &
  \multicolumn{2}{c}{55.0} &
  \multicolumn{2}{c}{80.0} &
  \multicolumn{2}{c}{9.3} &
  \multicolumn{2}{c}{\textbf{100.0}} &
  \multicolumn{2}{c}{92.7} &
  \multicolumn{2}{c}{59.8} \\
Llama3-8B &
  \multicolumn{2}{c}{88.3} &
  \multicolumn{2}{c}{93.2} &
  \multicolumn{2}{c}{21.4} &
  \multicolumn{2}{c}{60.0} &
  \multicolumn{2}{c}{83.3} &
  \multicolumn{2}{c}{23.3} &
  \multicolumn{2}{c}{\textbf{100.0}} &
  \multicolumn{2}{c}{96.7} &
  \multicolumn{2}{c}{{\underline{64.2}}} \\
SQLCoder-8B &
  \multicolumn{2}{c}{84.6} &
  \multicolumn{2}{c}{\textbf{99.6}} &
  \multicolumn{2}{c}{\textbf{77.8}} &
  \multicolumn{2}{c}{35.0} &
  \multicolumn{2}{c}{0.0} &
  \multicolumn{2}{c}{0.0} &
  \multicolumn{2}{c}{\textbf{100.0}} &
  \multicolumn{2}{c}{{\underline{98.0}}} &
  \multicolumn{2}{c}{59.7} \\
Codellama-7B &
  \multicolumn{2}{c}{\textbf{93.9}} &
  \multicolumn{2}{c}{16.5} &
  \multicolumn{2}{c}{4.3} &
  \multicolumn{2}{c}{\textbf{85.0}} &
  \multicolumn{2}{c}{{\underline{96.6}}} &
  \multicolumn{2}{c}{\textbf{66.3}} &
  \multicolumn{2}{c}{56.8} &
  \multicolumn{2}{c}{\textbf{100.0}} &
  \multicolumn{2}{c}{46.9} \\
Mistral-7B-v0.2 &
  \multicolumn{2}{c}{82.1} &
  \multicolumn{2}{c}{57.6} &
  \multicolumn{2}{c}{4.7} &
  \multicolumn{2}{c}{55.0} &
  \multicolumn{2}{c}{\textbf{100.0}} &
  \multicolumn{2}{c}{{\underline{50.0}}} &
  \multicolumn{2}{c}{79.1} &
  \multicolumn{2}{c}{77.5} &
  \multicolumn{2}{c}{55.3} \\ \bottomrule
\end{tabular}
\label{table:maintype}
\end{table*}

\paragraph{Clarification for Ambiguous Queries}

\begin{figure}[h]
  \centering
  \includegraphics[width=0.35\textwidth]{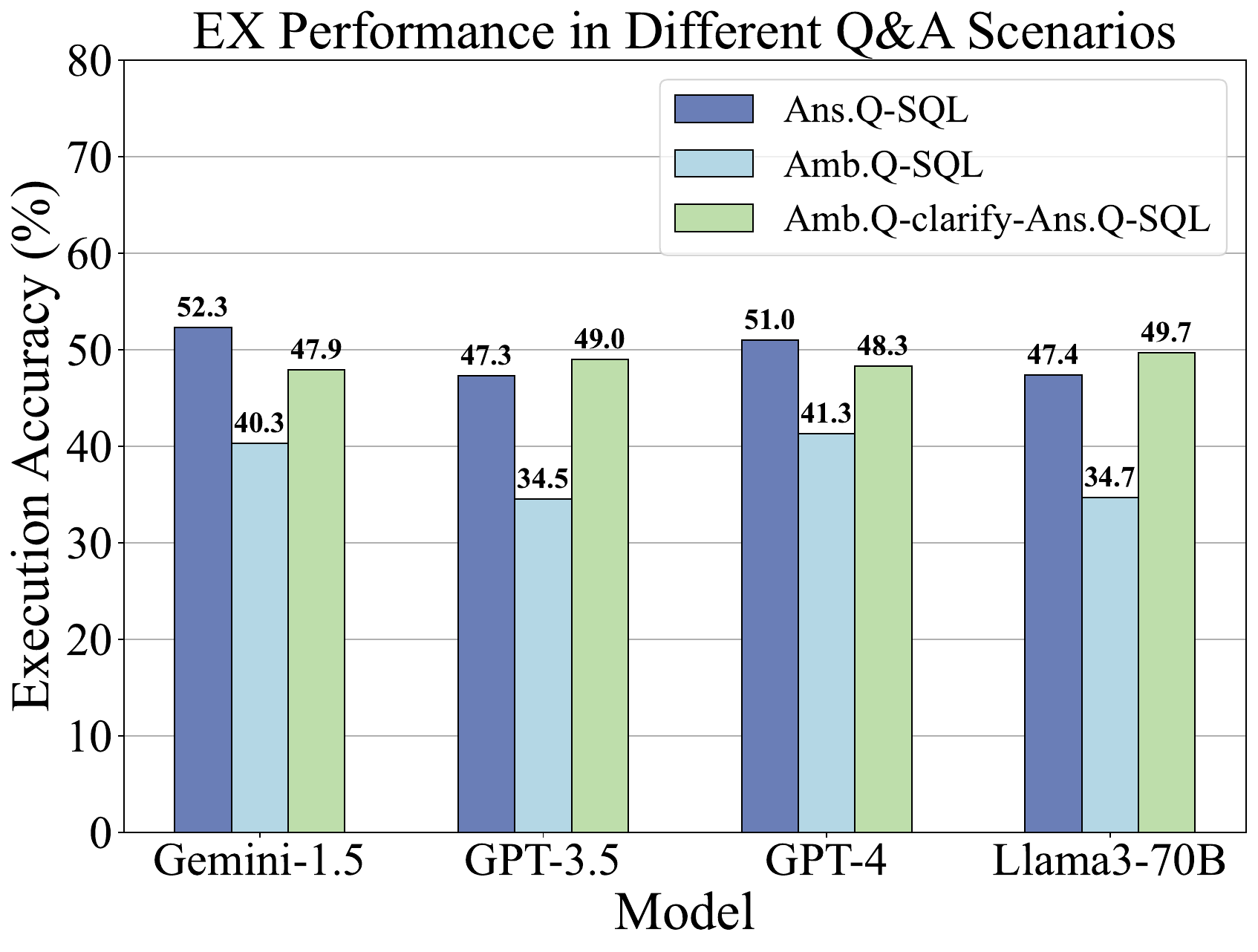}
  \caption{Performance analysis of model outputs on different Q\&A scenarios in the MMSQL test set}
  \label{fig:not_suitable_ans}
  \vspace{-4mm}
\end{figure}

As illustrated in Figure \ref{fig:not_suitable_ans}, several representative models exhibit a notable decline in execution accuracy when addressing ambiguous queries, compared to their performance with clear, answerable queries. For example, GPT-4 Turbo's execution accuracy decreases from 51.0\% with answerable queries to 41.3\% when handling ambiguous ones. Similarly, GPT-3.5 Turbo and Llama3-70B show significant declines, with accuracy falling from 47.3\% to 34.5\% and from 47.4\% to 34.7\%, respectively. However, execution accuracy improves significantly when models detect ambiguity and prompt users for clarification. For instance, when ambiguous queries are clarified, GPT-3.5 Turbo's query match accuracy rises from 34.5\% to 49.0\%. This suggests that generating SQL for ambiguous queries without clarification results in diminished performance. Therefore, it is crucial to generate SQL queries and effectively communicate ambiguities to users. By clarifying uncertainties and helping users understand them, models can assist in obtaining accurate information. Models can formulate more precise SQL queries by incorporating the clarification process across multiple interactions, enhancing text-to-SQL systems' usability and effectiveness.

\section{Our Solution: LLM-based Multi-agent Collaborative Framework}

\subsection{Schema Selector}
The Schema Selector is designed to identify the essential subset of a database schema necessary for answering a given question. By selectively focusing on the most relevant tables and columns, the Selector curtails the interference from extraneous data, enhancing operational accuracy. This retrieving capability is particularly pertinent in complex text-to-SQL systems, such as Business Intelligence (BI) platforms, which often interact with large databases containing numerous tables and columns. In our framework, the Selector is activated to choose tables and columns relevant to the interaction record solely when the schema size exceeds a specific threshold; otherwise, the full schema is employed. This adaptive mechanism ensures that our framework maintains efficiency and effectiveness across different database scales and complexities.

\subsection{Question Detector}
The Question Detector is responsible for detecting the question type and deciding on an appropriate answering strategy, whether to directly answer or attempt SQL generation, based on the subset of the database schema and the question-answer history. 

\begin{figure*}[h]
    \centering
    \includegraphics[width=0.85\linewidth]{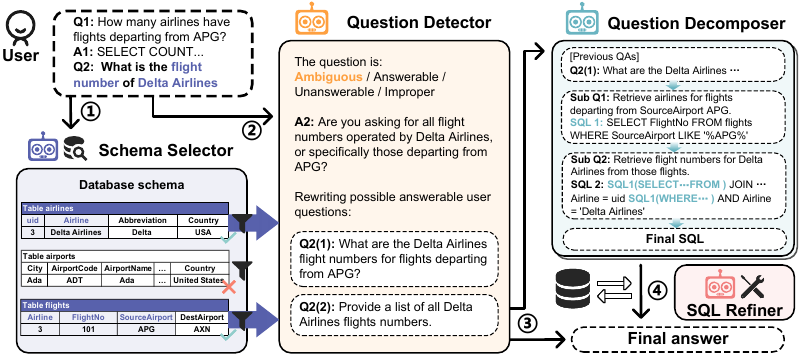}
    \caption{The overview of our multi-agent framework comprises four components: (i) the Schema Selector, which narrows down the database schema to focus on relevant tables, reducing noise from irrelevant data; (ii) the Question Detector, which determines its type and reformulates it if the question is deemed potentially ambiguous, potentially generating multiple possible rewrites; (iii) the Question Decomposer, which breaks down complex questions into simpler, manageable sub-questions and applies chain-of-thought reasoning to solve them progressively; and (iv) the SQL Refiner, which utilizes an external tool for SQL execution, gathers feedback, then refines faulty SQL queries.}
    \label{fig:multi-agent}
\end{figure*}

For answerable questions, the system proceeds to the Question Decomposer for further processing. In the case of ambiguous questions, where the query cannot be precisely mapped to the database schema, the system explains the ambiguity and attempts to rewrite it into answerable forms. For example, as shown in Figure \ref{fig:multi-agent}, the question "What is the flight number of Delta Airlines?" (Q2) could be interpreted in multiple ways: it might refer to flights departing from APG, as inferred from the previous context (Q1), or it could pertain to all flights in the database. Instead of rejecting such ambiguous queries, the system attempts to rewrite them into answerable questions and provides multiple potential answers in the final response.

For unanswerable questions, where the query is not ambiguous but cannot be answered due to data limitations or inappropriateness, the system informs the user of the reason. For improper questions, the LLM is tasked with providing a helpful response to assist the user as much as possible. In both scenarios, the system aims to guide the user towards obtaining a meaningful and helpful response as the final answer.

\subsection{Question Decomposer}
The Question Decomposer plays a critical role in the multi-agent framework by breaking down complex questions into simpler, manageable Sub-Questions \(\{Q_1, Q_2, \ldots, Q_{|Q|}\}\). For each Sub-Question \(Q_i\), it generates a corresponding sub-SQL \(\text{S}_j\) based on \(\text{S}_{<j}\), and the schema subset to address specific parts of the original query. This process involves applying chain-of-thought reasoning, allowing the system to progressively solve each sub-question. As shown in Fig. \ref{fig:multi-agent}, the Question Decomposer takes an ambiguous question, "What are the Delta Airlines flight numbers for flights departing from APG?", and decomposes it into sub-questions like "Retrieve airlines for flights from SourceAirport APG" and "Retrieve flight numbers for Delta Airlines from those flights." Each sub-question is then translated into a sub-SQL, collectively forming the final SQL output.

\subsection{SQL Refiner}
The SQL Refiner, as an integral auxiliary operation of the Question Decomposer, serves a critical role in ensuring the precision and reliability of the SQL queries generated by our multi-agent framework. This component is indispensable for the inspection and correction of the responses produced, particularly when addressing the complex demands of text-to-SQL tasks. Frequently utilized, the Refiner has demonstrated its effectiveness in enhancing the output to meet stringent accuracy requirements \cite{pourreza2024chasesqlmultipathreasoningpreference}. Drawing a parallel to its counterparts in software development, such as metaGPT \cite{hong2024metagptmetaprogrammingmultiagent} and ChatDev \cite{qian2024chatdevcommunicativeagentssoftware}, where intelligent agents are tasked with overall architectural design, code writing, and comprehensive testing, the SQL Refiner plays a similar role in the domain of database interactions. It adeptly adjusts SQL queries to accommodate various datasets, database schemas, and SQL generation styles, thereby ensuring the accuracy and effectiveness of the generated queries.

\begin{table}[h]
\centering
\caption{Performance metrics comparison on the MMSQL test set, demonstrating the improvements achieved by integrating the multi-agent framework with different models. Specific improvements are indicated in parentheses.}
\centering
\begin{tabular}{p{2.14cm}p{1.05cm}p{1.05cm}p{1.05cm}p{1.2cm}}
\toprule
\multicolumn{1}{c}{Model} & TDEX & EX   & RQS  & F1 Score \\ \hline
\textbf{GPT-4 Turbo}      &      &      &      &          \\
w/o Multi-Agent           & 67.0 & 70.0 & 5.8 & 68.2     \\
w/ Multi-Agent & 70.0(+3.0)          & 74.4(+4.4)          & 6.0(+0.2)          & 69.5(+1.3)           \\ \midrule
\textbf{Gemini-1.5 Flash} &      &      &      &          \\
w/o Multi-Agent           & 65.8 & 70.0 & 4.0 & 59.3     \\
w/ Multi-Agent & 69.4(+3.6)          & 73.7(+3.7)          & \textbf{7.0(+3.0)} & 70.7(+11.4)          \\ \midrule
\textbf{Llama3-70B}       &      &      &      &          \\
w/o Multi-Agent           & 62.8 & 66.4 & 3.9 & 59.8     \\
w/ Multi-Agent & \textbf{70.7(+7.9)} & \textbf{74.8(+8.4)} & 6.6(+2.7)          & \textbf{70.8(+11.0)} \\ \bottomrule
\end{tabular}
\footnotetext{The best values are marked in \textbf{bold}.}
\label{table:aftermulti}
\vspace{-4mm}
\end{table}

\section{Experiments}
\subsection{Experimental Setup}
We conducted a comprehensive experiment utilizing the MMSQL test suite to rigorously evaluate the efficacy of our multi-agent framework in managing intricate multi-turn text-to-SQL tasks. The experimental design was meticulously crafted to evaluate the framework's proficiency in producing valid SQL and coherent natural language responses, alongside its capacity to categorize diverse question types effectively. To quantitatively assess the quality of the natural language responses, we employed a suite of metrics, including the average RQS, EM, average F1 score across all categories, and TDEX, to provide a holistic evaluation of the models' performance.

We conduct experiments on three prominent models in Section \ref{sec:resultmmsql}: GPT-4 Turbo, Gemini-1.5 Flash, and Llama3-70b. These models were selected as our baselines to underscore the comparative effectiveness of our methodology. The objective of our experiments was to conduct a meticulous analysis elucidating how the incorporation of the multi-agent framework augments the performance of these models across a spectrum of evaluation metrics. For further details on the experimental implementation, refer to Appendix \ref{traindetail}.

\subsection{Overall Performance}

Table \ref{table:aftermulti} illustrates the performance of our method and baseline models on the MMSQL test set, highlighting the enhancements achieved by integrating our multi-agent framework with different models. The results demonstrate the benefits of our approach, with significant improvements across all metrics for each model. For example, the TDEX score increases from 62.8 to 70.7 and the F1 Score improves from 59.8 to 70.8 for Llama3-70B, showcasing the robustness of our framework in generating accurate SQL queries and natural language responses. These improvements underscore the effectiveness of our multi-agent framework in bolstering the capabilities of existing models to handle complex text-to-SQL tasks on the MMSQL benchmark.

\subsection{Ablation Study}
In Table \ref{table:multiablation}, we presents the findings from an ablation study on the MMSQL test set, examining the impact of removing individual components from the multi-agent framework when integrated with Gemini-1.5 Flash. The complete model achieves its best performance with scores of 69.4 in TDEX, 73.7 in EX, 7.05 in Average RQS, and an F1 Score of 70.7. The study reveals that the removal of any component results in a decrease in performance across these metrics.

\begin{table}[h]
\centering
\caption{Ablation study examining the impact of each component within the multi-agent framework on the Gemini-1.5 Flash's performance on the MMSQL test set.}
\begin{tabular}{lcccc}
\toprule
\multicolumn{1}{c}{Model}                     & \multicolumn{1}{c}{TDEX}  & \multicolumn{1}{c}{EX} & \multicolumn{1}{c}{Average RQS} & \multicolumn{1}{c}{F1 Score} \\ \midrule
\multicolumn{1}{l}{\textbf{Multi-Agent}} & \multicolumn{1}{c}{\textbf{69.4}} & \textbf{73.7}                       & \textbf{7.05}  & \textbf{70.7}\\
\multicolumn{1}{l}{w/o selector}              & 68.6                      & 73.4                     & 6.43 & 66.2  \\
w/o detector   & 66.1 & 70.8 & 5.57 & 64.4 \\
w/o decomposer & 68.6 & 73.1 & 6.33 & 68.1 \\
w/o refiner    & 67.6 & 70.5 & \textbf{7.05} & \textbf{70.7} \\
\bottomrule
\end{tabular}
\footnotetext{The best values are marked in \textbf{bold}.}
\label{table:multiablation}
\end{table}

Notably, the absence of the Detector results in the poorest performance with a sharp decline to 5.57 in Average RQS and 64.4 in F1 Score, indicating its pivotal role in selecting answering strategies. The Selector and Decomposer also contribute notably to the model's accuracy and response quality. The Refiner, also shows a notable impact when removed, particularly affecting the EX, dropping to 70.5, suggesting its role in correcting errors is indispensable for ensuring the accuracy of the final SQL queries generated.

\section{Discussion}\label{sec5}

In this study, we introduced the MMSQL test suite, a benchmark for evaluating LLMs on complex multi-turn and multi-type text-to-SQL tasks. Evaluating eight LLMs, we found significant challenges with MMSQL's diverse question types, particularly ambiguous or unanswerable queries. This underscores the need for models with better intent understanding and uncertainty management. Notably, open-source models like the Llama3 series matched or surpassed closed-source models on several metrics, suggesting avenues for developing more reliable text-to-SQL systems. We also developed a multi-agent framework for handling multi-type text-to-SQL problems, unlike previous approaches that abstained when facing ambiguous queries \cite{lee2024trustsql,wang2023know}. By refining the existing multi-agent framework \cite{wang2024macsqlmultiagentcollaborativeframework} and adding a Question Detector Agent, our framework resolves ambiguities by requesting user clarifications, improving query handling and user interaction.

Despite these advancements, limitations remain. The performance of open-source LLMs could be improved with larger models. Our multi-agent framework currently relies on closed-source LLMs. Exploring open-source, domain-specific LLMs and further annotating the MMSQL training set could enhance our framework's effectiveness \cite{wang2024llm}, providing tailored training data for each agent and allowing targeted fine-tuning for better results \cite{lu2024finetuninglargelanguagemodels}. Additionally, integrating domain-specific knowledge bases could improve the model's ability to handle queries in specialized  fields.

\section{Conclusion}\label{sec6}


Current text-to-SQL evaluation methods often emphasize SQL complexity, yet struggle with conversational dynamics, limiting their effectiveness. To address these challenges, we introduce the MMSQL test suite and a novel multi-agent framework. MMSQL offers a structured evaluation environment with diverse, multi-turn question types, highlighting limitations in existing LLMs. Our multi-agent framework effectively navigates these complexities, enhancing the robustness and usability of text-to-SQL systems. This work advances data accessibility and provides practical insights for achieving high performance in this domain.

\bibliographystyle{IEEEtran}
\bibliography{IEEEabrv}{}

\begin{appendices}

\section{Implementation Details}
\label{traindetail}
To ensure reproducibility, we applied greedy decoding strategies during both inference and evaluation. The experiments were conducted on a server equipped with an Intel Xeon Gold 6133 CPU and four NVIDIA A800 80GB PCIe GPUs.

\begin{table*}[ht]
\caption{Evaluation Criteria for Response Quality}
\centering
\begin{tabular}{ccp{10cm}}
\toprule
\textbf{Criterion} & \textbf{Score} & \textbf{Description} \\ \midrule
\multirow{3}{*}{Relevance} & 0 & The response is completely irrelevant. \\ \cline{2-3} 
 & 1 & The response is partially relevant but misses key details. \\ \cline{2-3} 
 & 2 & The response is fully relevant and adequately addresses the question. \\ \midrule
\multirow{3}{*}{Clarity} & 0 & The response is incomprehensible. \\ \cline{2-3} 
 & 1 & The response is mostly clear with minor ambiguities. \\ \cline{2-3} 
 & 2 & The response is very clear and easy to understand. \\ \midrule
\multirow{3}{*}{Completeness} & 0 & The response does not address the question at all. \\ \cline{2-3} 
 & 1 & The response covers most aspects of the question but lacks some details. \\ \cline{2-3} 
 & 2 & The response thoroughly addresses all aspects of the question. \\ \midrule
\multirow{3}{*}{Accuracy} & 0 & The response contains factually incorrect information. \\ \cline{2-3} 
 & 1 & The response is partially accurate with some errors. \\ \cline{2-3} 
 & 2 & The response is completely accurate. \\ \midrule
\multirow{3}{*}{Utility} & 0 & The response does not meet the user's needs or explain the context of the question. \\ \cline{2-3} 
 & 1 & The response somewhat meets the user's needs and provides partial explanations. \\ \cline{2-3} 
 & 2 & The response excellently meets the user's needs and clearly explains the context or ambiguity of the question. \\ \bottomrule
\end{tabular}

\label{table:criteria}
\end{table*}

\section{Design and Effectiveness Study of QDA-SQL for Data Generation}
\label{QDA-SQL}

\begin{figure}[h]
    \centering
    \includegraphics[width=0.9\linewidth]{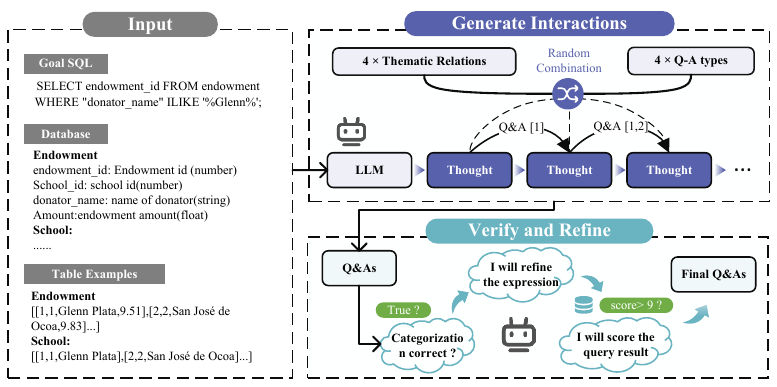}
    \caption{Overview of QDA-SQL processes.}
    \label{fig:qda}
\end{figure}

In this study, we use QDA-SQL to expand the original datasets by inputting samples from SParC and CoSQL, while adhering to MMSQL's classification and formatting standards. As depicted in Figure \ref{fig:qda}, QDA-SQL first generates multi-turn question-and-answer pairs in the Generate Interactions phase. During this phase, each Q\&A pair is assigned a randomly selected thematic relation (as defined by \cite{yu2019sparc}) and question type, referencing the existing dialogue history to maintain contextual relevance. In the subsequent Verify and Refine phase, Gemini Pro optimizes, scores, and filters these pairs to ensure alignment with our question type requirements and maintain high quality.

To evaluate the quality of the data generated by QDA-SQL, A manual review of the automatically filtered results was conducted to evaluate the quality of the data generated by QDA-SQL, focusing on both classification accuracy and the quality of the annotated natural language question answering. A comparison of samples before and after the Verify and Refine process revealed that Gemini Pro successfully identified 94\% (47 out of 50) of the misclassified samples. This finding underscores the effectiveness of the automatic filtering process in enhancing data quality and supporting high classification accuracy.

\begin{figure}[h]
    \centering
    \includegraphics[width=0.8\linewidth]{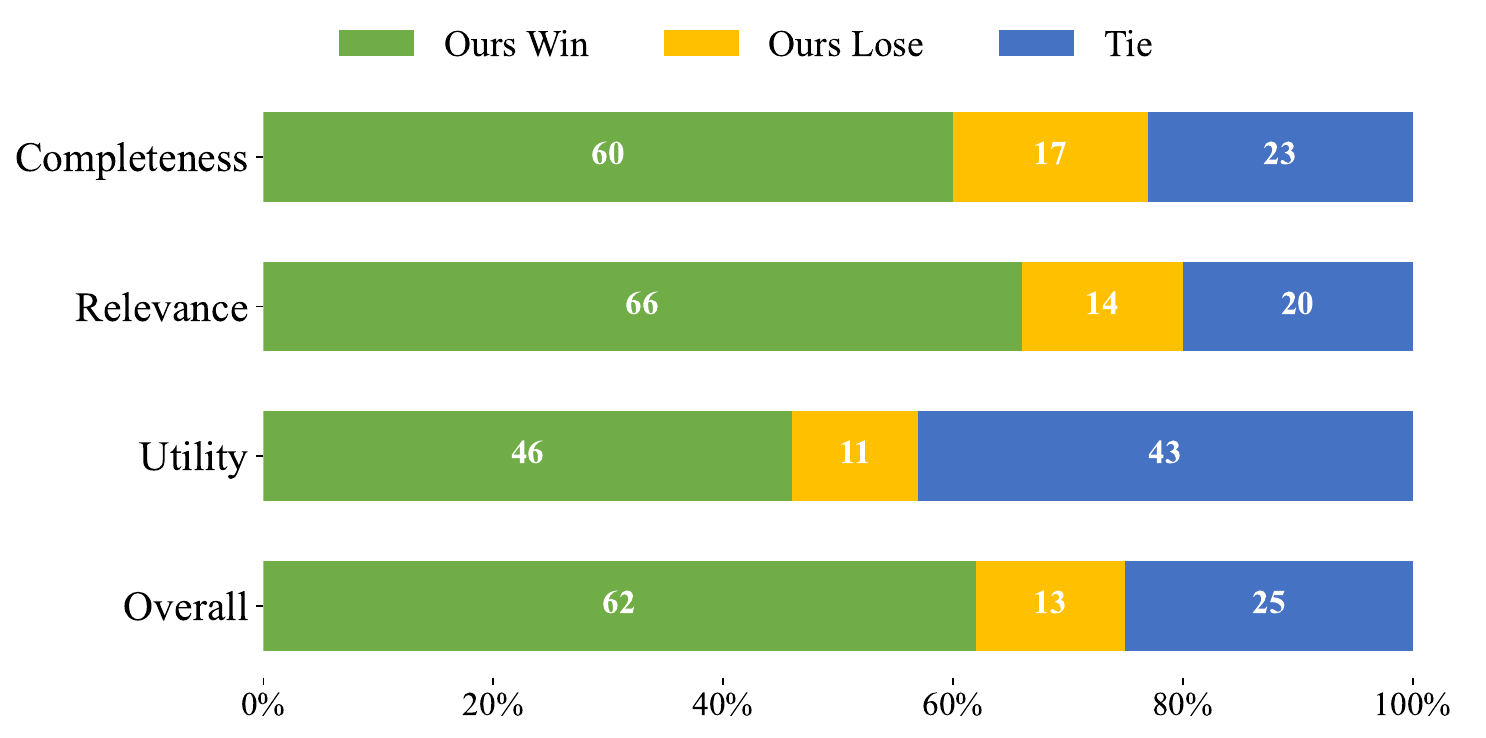}
    \caption{Pairwise comparison of original and QDA-SQL enhanced dataset annotation quality across different criteria.}
    \label{fig:data_comparison_results}
    \vspace{-4mm}
\end{figure}

Additionally, an automatic evaluation framework based on GPT-4 was implemented, the effectiveness of which has been confirmed by numerous studies \cite{NEURIPS2023_91f18a12, xu2023wizardlmempoweringlargelanguage}. Critical dimensions include completeness, relevance, and utility in answering the annotated natural language question. To address potential order bias \cite{iourovitski2024gradescorequantifyingllm}, the placement of the original dataset and the enhanced dataset was alternated in pairwise comparisons, positioning the enhanced dataset first for odd IDs and second for even IDs. As illustrated in Figure \ref{fig:data_comparison_results}, the enhanced dataset demonstrated superior performance, with an overall assessment indicating that 62\% of the QDA-SQL enhanced dataset is considered superior to the manually annotated original SParC and CoSQL samples.

\section{LLM-Assisted RQS Metric Evaluation and Cost}
\label{rqsp}

We used GPT-4o-mini to evaluate natural language responses across five dimensions: Utility, Accuracy, Completeness, Clarity, and Relevance (see Table \ref{table:criteria}). Only the RQS metric in our MMSQL testing employed the OpenAI GPT-4o-mini API through LLM-assisted rating. We scored each of four metrics from 0 to 2, with a maximum total score of 10. This process incurred API costs—currently \$0.15 per million input tokens and \$0.60 per million output tokens for GPT-4o-mini. Scoring our 149 test rounds cost approximately \$0.036, significantly less than manual scoring, which relies heavily on human labor.

\end{appendices}

\end{document}